\documentclass{article}

\usepackage[preprint]{neurips_2021}

\usepackage[utf8]{inputenc} 
\usepackage[T1]{fontenc}    
\usepackage{hyperref}       
\usepackage{url}            
\usepackage{booktabs}       
\usepackage{amsfonts}       
\usepackage{nicefrac}       
\usepackage{microtype}      
\usepackage{xcolor}         
\usepackage{amsmath}
\usepackage{amssymb}
\usepackage{soul}
\usepackage{algorithm}
\usepackage{algpseudocode}
\usepackage{graphicx}
\usepackage{caption}
\usepackage{subcaption}
\usepackage{multirow}

\title{Natural Gradient Methods: Perspectives, Efficient-Scalable Approximations, and Analysis}

\author{%
  Rajesh Shrestha\\
  School of Electrical Engineering and Computer Science\\
  Oregon State University\\
  Corvallis, OR 97331 \\
  \texttt{shresthr@oregonstate.edu}
}

\begin{document}
\maketitle
\begin{abstract}
Natural Gradient Descent, a second-degree optimization method motivated by the information geometry, makes use of the Fisher Information Matrix instead of the Hessian which is typically used. However, in many cases, the Fisher Information Matrix is equivalent to the Generalized Gauss-Newton Method, that both approximate the Hessian. It is an appealing method to be used as an alternative to stochastic gradient descent, potentially leading to faster convergence. However, being a second-order method makes it infeasible to be used directly in problems with a huge number of parameters and data. This is evident from the community of deep learning sticking with the stochastic gradient descent method since the beginning. In this paper, we look at the different perspectives on the natural gradient method, study the current developments on its efficient-scalable empirical approximations, and finally examine their performance with extensive experiments.
\end{abstract}
\section{Introduction}
The gradient descent method \cite{10.1214/aoms/1177729392,10.1214/aoms/1177729586} has been very popular in non-linear optimization including the field of machine learning and deep learning. The stochastic version i.e. Stochastic Gradient Descent method (SGD) is a simple and light method that scales very well to train big deep learning models on a vast amount of data. Almost all the work in the field of deep learning has used SGD in some form for training the model. This has been used in all domains including computer vision, reinforcement learning, natural language processing, etc. in tasks like face recognition \cite{hu2015face}, object recognition \cite{eitel2015multimodal}, control \cite{huang2019adaptive} and much more. This first-order optimization method utilizes the empirical estimation of the gradient in the parameter space. There have been multiple variations to this method like Nesterov accelerated gradient \cite{nesterov1983method}, SGD with momentum \cite{QIAN1999145}, RMS prop \cite{hinton2012neural}, Adam \cite{kingma2014adam}, etc. that improves its stability and convergence. However, the performance of this method is limited being a first-order method. There has been much interest in using the second-order optimization method to improve the convergence and reduce the training time \cite{01b8a4abbaba43b48ce43466318a9927, grosse2016kronecker, osawa2020scalable}, and the natural gradient descent method \cite{grosse2016kronecker, osawa2020scalable, zhang2019fast} has grasped the attention of the community.

As the name implies, the natural gradient method \cite{10.1162/089976698300017746} uses the natural gradient instead of the standard gradient which is defined as the product of the inverse of the Fisher Information matrix(FIM) and the gradient. This method has been motivated to take the steepest descent in the space of realizable distribution rather than the space of parameters, where the "Riemannian metric" is used for computing distance in the distribution \cite{amari2000methods}. This metric approximates the square root of KL divergence locally and is only dependent on the distribution itself, not the parameterization. From this perspective of natural gradient, this method is invariant to any smooth and invertible reparameterization of the model \cite{ollivier2013riemannian, martens2020new}. 

The natural gradient method, being closely related to the Gauss-Newton method, seems to require a much less number of iterations than SGD making it a potential and appealing method. But this doesn't always translate to faster convergence in practice. \cite{10.1162/089976698300017746} argued that this is contingent on how well the cost function being optimized is well approximated by a convex quadratic. The training of neural networks involves optimizing the non-convex and potentially non-smooth loss functions, making it difficult to prove any form of convergence guarantee or bound. In addition, the large size of the Fisher Information Matrix in the case of highly parameterized neural networks makes it infeasible to be used. This problem can be mitigated through approximation of FIM such that the computation, storage, and inversion of this matrix is done efficiently \cite{grosse2016kronecker, george2018fast, goldfarb2020practical, soori2021tengrad}. In this paper, we study the theoretical definition, and interpretation along with various efficient-scalable versions of the natural gradient method and compare them with SGD to study its properties through empirical study.

In the following section, we first define the Fisher Information matrix and derive its relationship with the Hessian matrix. After it, we will define natural gradient and discuss its geometric interpretation leading to the derivation of natural gradient methods. Finally, we will discuss some of the approximate natural gradient methods and analyze them through experiments.
\section{Background\protect\footnote{The mathematical deviations in this section are adapted from the works of \cite{martens2020new, kristiadi}}}
\newcommand{\xsample}{x}
\newcommand{\likelihood}{L(\theta|\xsample)}
\newcommand{\loglikelihood}{l(\theta|\xsample)}
\newcommand{\probability}{p(\xsample|\theta)}
\newcommand{\expectation}{\mathbb{E}}
\newcommand{\scorefunction}{s(\theta|\xsample)}
\newcommand{\hessian}{\mathbf{H}}
\newcommand{\jacobian}{\mathbf{J}}
\newcommand{\fim}{\mathbf{F}(\theta)}

\subsection{Fisher Information Matrix(FIM)}
\paragraph{Definition:}
Suppose we have a model parameterized by $\theta$ that models the distribution $p(\xsample|\theta)$. Then, the likelihood of the parameter $\theta$ is given by equation \ref{eqn: likelihood} and similarly, the loglikelihood is given by equation \ref{eqn: loglikelihood}.
\begin{align}
    \likelihood &= \probability \label{eqn: likelihood}\\
    \loglikelihood &= log(\likelihood) \label{eqn: loglikelihood} \\
    &=log(\probability) \nonumber
\end{align}
The parameters are usually estimated by maximizing this loglikelihood function in the frequentist approach. Hence, the negative of this loglikelihood is often referred to as loss/cost in training. The necessary condition for this estimated $\theta$ to be an optimal one is that this $\theta$ point should correspond to one of the stationary points in the likelihood function. Hence, we use the gradient of likelihood function defined as a score function (equation \ref{eqn: score function}) to analyze the estimated value.
\begin{equation}
    \scorefunction = \nabla_{\theta}\loglikelihood \label{eqn: score function}
\end{equation}
\emph{Then, the expected value of this score function is 0} (Appendix \ref{appendix proof: expectation of score function}).
\begin{equation}
    \therefore \expectation_{\probability}[\scorefunction] = 0 \label{eqn: expectation of score function}
\end{equation}
\emph{Then, the fisher information matrix (FIM) is basically the expected measure of uncertainty in the $\mathbf{\scorefunction}$ around the expected value i.e. covariance matrix of the score function (equation \ref{eqn: FIM primary definition}) and contains the curvature information.} For simplicity, we denote the score function by only $s$ below.
\begin{align}
    \fim &= \expectation_{\probability}[(s-\expectation_{\probability}[s])(s - \expectation_{\probability}[s])^T] \label{eqn: FIM primary definition}\\
    &= \expectation_{\probability}[\nabla_{\theta}\loglikelihood. \nabla_{\theta}\loglikelihood^T] \label{eqn: FIM final definition}
\end{align}
Usually, it's not feasible to compute the expectation in equation \ref{eqn: FIM final definition}, especially in the neural networks as the distribution is very complicated and most likely very high dimensional. In such cases, the expectation is intractable. It's common for it to be estimated using the data distribution $q(\xsample)$ instead. Given the data sample of size $N$, the empirical estimation of the FIM is given by
\begin{equation}
    \fim = \frac{1}{N} \sum_{i=1}^{N} \nabla_{\theta}logp(\xsample_i|\theta) \nabla_{\theta}logp(\xsample_i|\theta)^T \label{eqn: empirical FIM}
\end{equation}

\paragraph{Relation between FIM and Hessian:}
\emph{The Fisher Information Matrix (FIM) is equal to the negative of the expected value of the hessian of loglikelihood as shown in equation \ref{eqn: fim and hessian}(Appendix \ref{appendix proof: relationship between Hessian and FIM}) and measures the curvature information which is further explained in section \ref{section: KL and FIM} }
\begin{equation}
    \therefore \fim = - \expectation_{\probability}[\hessian(\loglikelihood)] \label{eqn: fim and hessian}
\end{equation}

\newcommand{\loss}{\mathcal{L}}
\newcommand{\sampledata}{\mathbb{X}}
\subsection{Natural Gradient}
Let a model parameterized by $\theta$ as before that models the distribution $\probability$. Let $\sampledata$ denote the sample data of size $N$ and $\mathcal{L(\theta|\sampledata)}$ denotes the loss function i.e. negative loglikelihood function (equation \ref{eqn: loss function}).

Simply, the natural gradient is the product of the inverse of FIM and the gradient  as shown in equation \ref{eqn: natural gradient}. In the following subsections, we examine the geometric interpretation of this natural gradient and show how this quantity is derived \cite{martens2020new}.
\begin{equation}
    \tilde{\nabla}_{\theta} = F(\theta)^{-1}\nabla_{\theta}\loss \label{eqn: natural gradient}
\end{equation}
\begin{equation}
    \mathcal{L(\theta|\sampledata)} = -L(\theta|\sampledata) \label{eqn: loss function}
\end{equation}
\subsubsection{Distribution Space:}
The update equation for the original gradient descent uses the gradient of the loss function $\loss$ to update in the parameter space.
\begin{equation}
    \theta^+ = \theta - \alpha\nabla_{\theta}\loss(\theta|\sampledata) \text{ ,where $\alpha$ is learning rate}
\end{equation}
This method selects the direction on the $\theta$ space so that the loss $\loss$ decreases the most i.e. highest reduction in loss with a unit change in the parameter $\theta$. Mathematically,
\begin{equation}
    -\frac{\nabla_{\theta}\loss}{||\nabla_{\theta}\loss||} = \lim_{\epsilon \rightarrow 0}\frac{1}{\epsilon}\arg\min_{t: ||t||\leq \epsilon}\loss(\theta+t) \label{eqn: steepest descent SGD}
\end{equation}
The equation \ref{eqn: steepest descent SGD} shows that the steepest direction is the one that is within $\epsilon$ proximity such that $\loss(\theta+t)$ is the minimum obtainable with the constraint. We can see that the neighborhood is defined by $||t||$ which is dependent on the euclidean space of parameters $\theta$, as a result, this method is dependent on the euclidean geometry of parameter space.

Let's suppose, the actual underlying distribution of the data generation process is $q(x)$ which isn't available to us. Then our aim is to learn $\theta$ so that the distance between the two distributions -- distribution with learned model $\probability$ and the underlying distribution $q(x)$ --  is smaller. \emph{Hence, it is more logical and sensible to take a step in the direction of descent in this distance in the model's distribution space instead of euclidean parameter space}. \emph{Intuitively, taking even a small step in parameter space could result in a drastic change in the distribution so a step in the realizable distribution space should be taken instead.} One form of measure of distance is the KL divergence.

\paragraph{Relation between KL divergence and Fisher Information Matrix:} \label{section: KL and FIM}
\emph{FIM is the hessian of KL-divergence of two distributions $p(x|\theta)$ and $p(x|\theta')$ evaluated at $\theta'=\theta$(Appendix \ref{appendix proof: FIM and KL divergence}). FIM defines the curvature of the distribution space using the KL divergence as the metric}.
\begin{equation}
    \fim=\nabla_{\theta'}^2\text{KL}(p(x|\theta)||p(x|\theta'))|_{\theta'=\theta}
\end{equation}

\subsection{Natural Gradient Descent(NGD): Steepest Descent in the distribution space}
Let's approximate the KL divergence between two distributions with parameters $\theta$ and $\theta'$ such that $t=\theta'-\theta$. In the following for simplicity, we use the notation $p_\theta$ to denote $p(x|\theta)$. Using second-order Taylor expansion,
\begin{align}
    \text{KL}(p_\theta||p_{\theta'}) &\approx \text{KL}(p_{\theta}||p_{\theta'})|_{\theta'=\theta} + t(\nabla_{\theta'}\text{KL}(p_{\theta}||p_{\theta'}))|_{\theta'=\theta}+\frac{1}{2}t^T(\nabla_{\theta'}^2\text{KL}(p_{\theta}||p_{\theta'}))|_{\theta'=\theta}t \nonumber \\
    &= 0 - t \expectation_{p_\theta}(\nabla_{\theta'}log(p_\theta)) + \frac{1}{2}t^TF(\theta)t \nonumber \\
    &= 0- 0 + \frac{1}{2}t^TF(\theta)t \nonumber \\
    \therefore \text{KL}(p_\theta||p_{\theta'}) &\approx  \frac{1}{2}t^TF(\theta)t
\end{align}
Now, let's formalize the problem of finding the steepest descent in the distribution space. Let $t^*$ be the direction corresponding to the steepest descent similar to equation \ref{eqn: steepest descent SGD}, then,
\begin{equation}
    t^* = \arg\min_{t \text{ s.t. KL}(p_\theta||p_{\theta+t})<=\epsilon^2} \loss(\theta+t)
\end{equation}
Using Lagrange Multiplier, we get
\begin{align}
    t^*, \lambda^* &= \arg\max_\lambda  \arg\min_t \loss(\theta+t) + \lambda(KL(p_\theta||p_{\theta+t})-\epsilon^2) \\
    & \approx \arg\max_\lambda  \arg\min_t \loss(\theta) + t\nabla_\theta\loss(\theta) + \lambda\left(\frac{1}{2}t^T F(\theta)t-\epsilon^2\right)
\end{align}
Taking the first derivatives we get,
\begin{align}
    &\nabla_{\theta}\loss(\theta)+\lambda F(\theta)t = 0 \nonumber \\
    &t^* = -\frac{1}{\lambda^*}F(\theta)^{-1}\nabla_{\theta}\loss(\theta)
\end{align}
By dual feasibility, the value of $\lambda^* >=0$, hence the direction of steepest descent is given by $-F(\theta)^{-1}\nabla_{\theta}\loss(\theta)$ which is same as the negative of quantity that we defined as the natural gradient in equation \ref{eqn: natural gradient}. The constant positive term can be absorbed in the learning rate $\alpha$.

\begin{algorithm}
    \caption{Natural Gradient Descent Method}
    \begin{algorithmic}[1]
        \State \textbf{Input}: data $\sampledata$, learning rate $\alpha$
        \State Initialize model parameters $\theta$
        \While{not convergence in $\theta$}
        \State Sample data and forward pass in the model
        \State Compute the loss $\loss$ for the passed data
        \State Compute the gradient $\nabla_{\theta}\loss$
        \State Compute FIM or its empirical approximation: \\
        \;\;\;\;\;\;\;\;\;\;\;$\fim = \frac{1}{N} \sum_{i=1}^{N} \nabla_{\theta}logp(\xsample_i|\theta) \nabla_{\theta}logp(\xsample_i|\theta)^T$
        \State Compute the natural gradient: $\tilde{\nabla}_{\theta} =  F^{-1}\nabla_{\theta}\loss(\theta)$
        \State Update Model parameter: $\theta^+ = \theta - \alpha\tilde{\nabla}_{\theta}$
        \EndWhile
    \end{algorithmic}
\end{algorithm}

\section{Efficient-Scalable Approximations}
Despite having strong theoretical significance and potential of natural gradient, this method in its original form is limited in the scope of models to which it can be applied. Most of the neural networks that are used in practice have a very high number of parameters that makes the use of this method in them infeasible as a very high dimensional FIM matrix needs to be computed, stored, and inverted too. The number of elements in FIM scales quadratically to the number of parameters in a model. This is the main reason why simple first-order gradient methods have been popular over these better second-degree methods. 

\begin{figure}
    \centering
    \includegraphics[width=\textwidth]{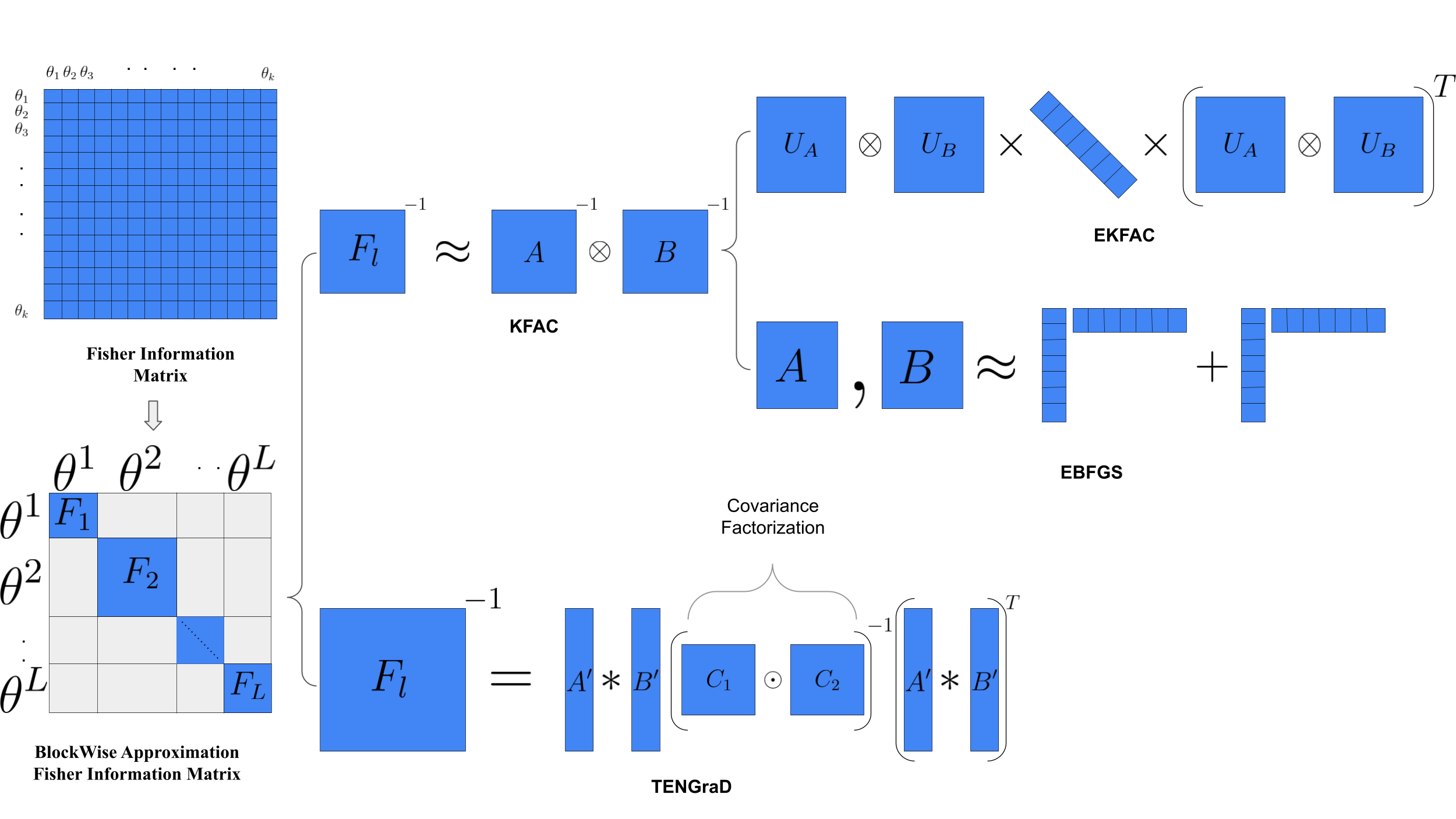}
    \caption{Approximations of FIM in natural gradient method \cite{soori2021tengrad}}
    \label{fig: approximations of FIM}
\end{figure}

Suppose we have a deep neural network $f$ with $L$ layers  parameterized by $\theta$ with $p$ number of parameters in total. Let $\theta^l$ be the parameters corresponding to each layer of the network. Let $d_i^l$ be the input size and $d_o^l$ be the output size of the $l^{th}$ layer. With the use of exact FIM, the storage required is $O(p^2)$ and the computation of the inverse of FIM for natural gradient (equation \ref{eqn: natural gradient}) requires $O(p^3)$. To mitigate this problem, most of the natural gradient methods approximate FIM using a block-diagonal matrix such that the elements corresponding to cross layers are 0 as shown in figure \ref{fig: approximations of FIM}. The problem reduces to finding the FIM corresponding to each layer i.e. $F_l$ for $l=1,....., L$. 

Even with this block-diagonal approximation, the requirement in storage and computation is still high for layers that have a high number of parameters, for instance, a fully connected layer with high input and output dimensions, the requirement in storage and computation is high. To mitigate this issue, the inverse of block FIM is further approximated using efficient and smaller matrices. KFAC \cite{martens2015optimizing, grosse2016kronecker} uses the Kronecker product of two small matrices to approximate the inverse of FIM for a layer. The inverse can still be expensive for wide layers so EKFAC \cite{george2018fast} improves this approximation by rescaling the Kronecker factors using a diagonal matrix which is obtained by using svd. In contrast, EBFGS approximates the inverse of the Kronecker factors using low-rank BFGS updates. Among all the development, a recent method TENGraD \cite{soori2021tengrad} performs better in terms of approximation accuracy, time efficiency, and storage requirement. This method performs exact FIM block inversion by using computationally efficient covariance factorization and reusing the intermediate vectors and matrices. Further details on TENGraD can be found in Appendix \ref{tengrad description}. In the following sections, we perform experiments using these methods and analyze the results obtained.

\section{Experiments}
Apart from the theoretical interpretation and claims, I performed experiments to see how much those things hold in practice and how the different approximations compare to the gradient-based method SGD. The experiment was performed in the deep neural network(DNN) in different settings and using different datasets. The set of experiments performed here can be extended to other architectures of neural network including but not limited to Convolutional Neural Networks (CNN). While working with the deep models in this experiment, TENGraD is mostly used as a natural gradient method as other methods aren't scalable to models with a huge number of parameters. In addition, the TENGraD method is equivalent to a block-wise natural gradient but an efficient one. My analysis targets three properties that are important to evaluate any optimization method: \emph{Convergence}, \emph{Performance}, \emph{Stability}, and \emph{Scalability}.

\subsection{Convergence} \label{subsection: convergence}
\begin{figure}[h!]
\centering
\begin{subfigure}[b]{0.48\textwidth}
    \centering
    \includegraphics[width=\textwidth]{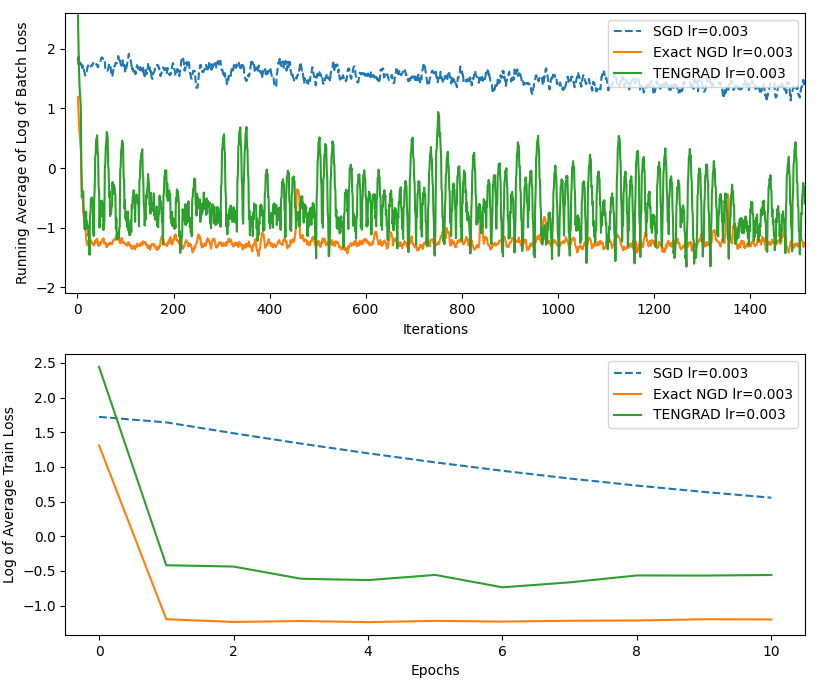}
    \caption{Same learning rate($\alpha =3*10^{-3}$)}
    \label{fig: figure 1 same lr}
\end{subfigure}
\hfill
\begin{subfigure}[b]{0.48\textwidth}
    \centering
    \includegraphics[width=\textwidth]{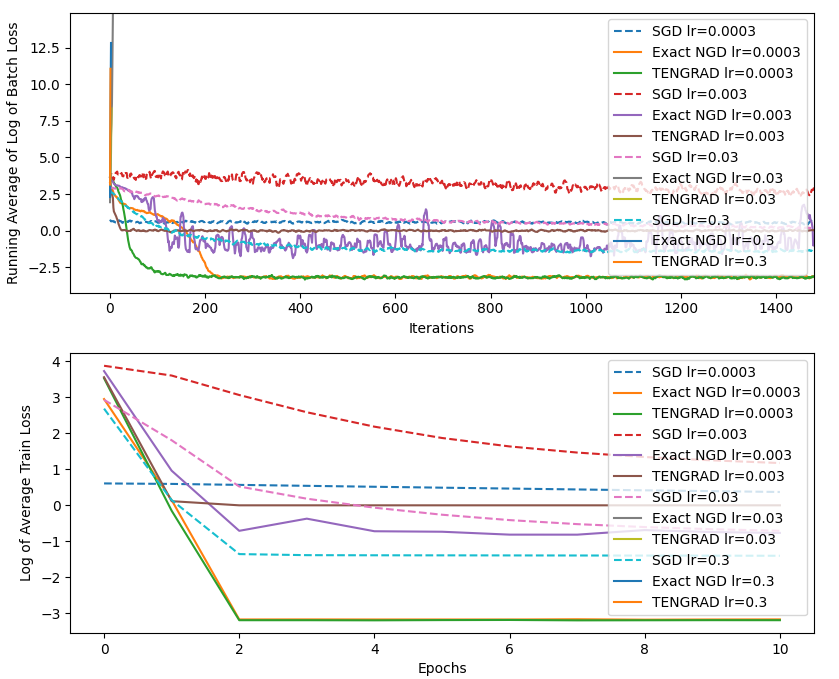}
    \caption{Grid search on learning rate($\alpha$)}
    \label{fig: figure 2 grid lr}
\end{subfigure}
    \caption{Convergence in DNN using weather dataset for all methods(with one hidden layer of size 4) and batch size=128}
    \label{fig:  figure 12 DDN(4) weather}
\end{figure}

Being the second-degree method, the natural gradient methods (both exact NGD and TENGraD) do often converge very faster in practice, even with the empirical FIM matrix. Exact NGD method seems to perform better than the approximation-based method which is expected. From figure \ref{fig:  figure 12 DDN(4) weather}, it's clear that the natural gradient method converges faster than the gradient-based method(SGD). This is not only due to the effective learning rate $\alpha$ induced by the Fisher Information Matrix(FIM) which can be seen with the grid search for lr in figure \ref{fig: figure 2 grid lr}. Even with using the best $\alpha$s, the natural gradient method(NGD) still performs better than SGD. We can still see similar behavior in deep and wide networks given that the algorithms are made stable for these models.

\begin{figure} [h!]
\centering
\begin{subfigure}[b]{0.48\textwidth}
    \centering
    \includegraphics[width=\textwidth]{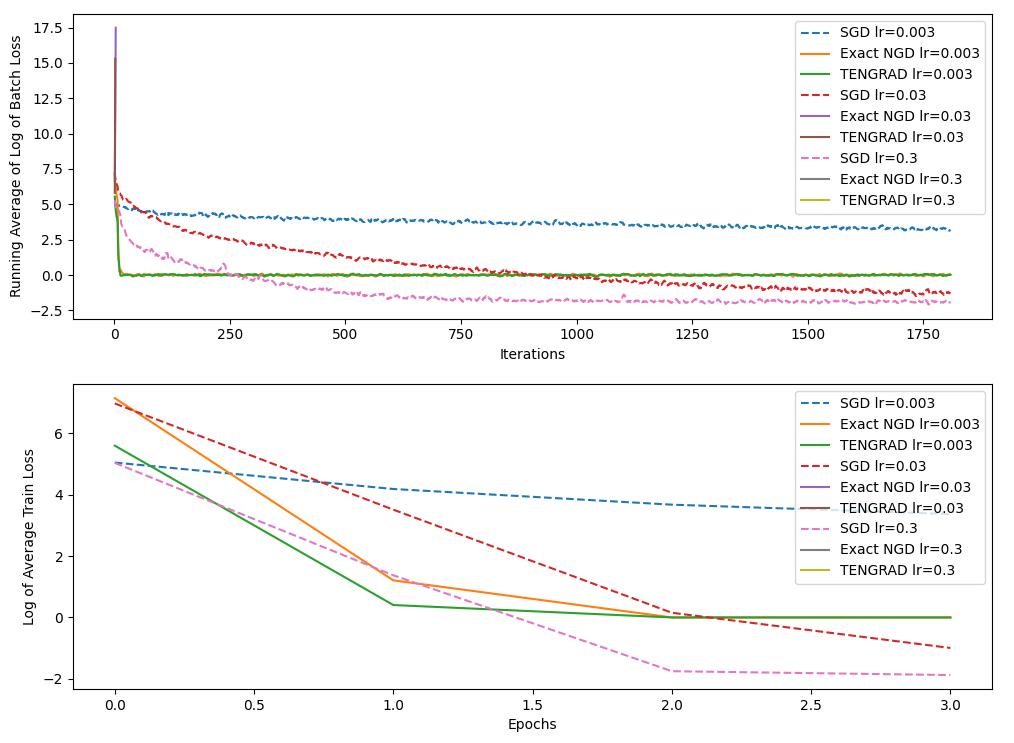}
    \caption{Hidden layers=[32, 16]}
    \label{fig: figure 3 wide dnn}
\end{subfigure}
\hfill
\begin{subfigure}[b]{0.48\textwidth}
    \centering
    \includegraphics[width=\textwidth]{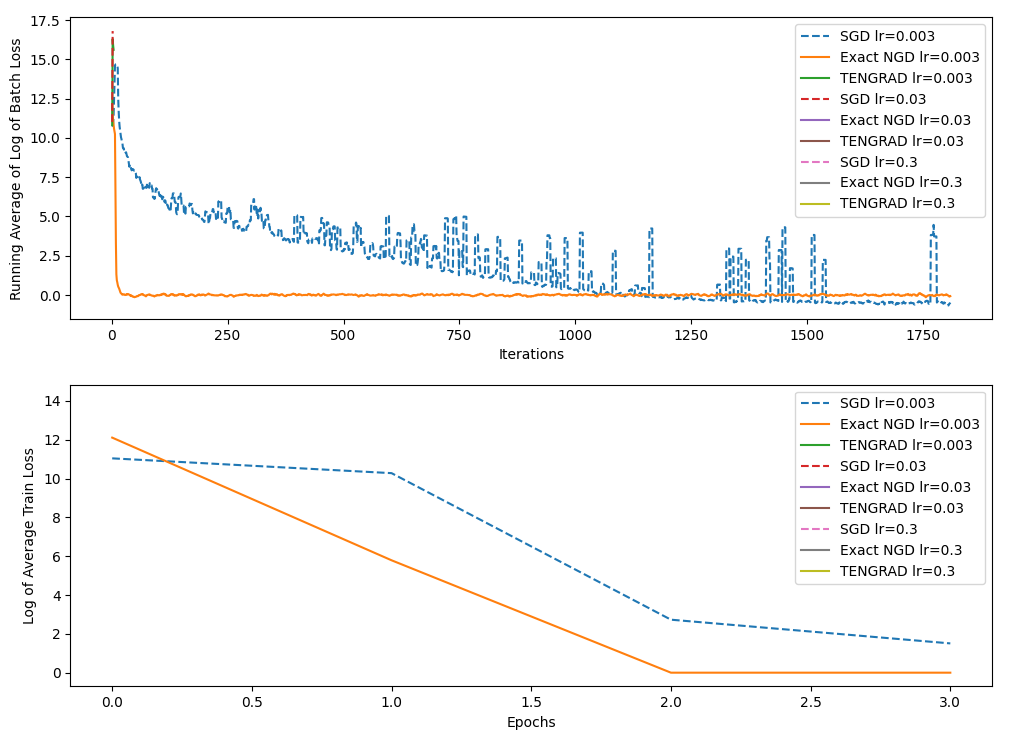}
    \caption{Hidden layers=[32, 32, 32, 16]}
    \label{fig: figure 4 deep and wide dnn}
\end{subfigure}
    \caption{Convergence in Deep and/or Wide DNN using grid lr search and  weather dataset with batch size=128}
    \label{fig:  figure 34 Deep and/or Wide DDN(4) weather}
\end{figure}

This convergence property seems to hold for deep models with wide layers too (figure \ref{fig:  figure 34 Deep and/or Wide DDN(4) weather}).

Both SGD and NGD use a mini-batch of data to estimate the gradient and hessian locally. This estimate will be noisy on using a sub-sample of data and will impact the convergence of both methods. However, NGD uses both first-order and second-order derivatives estimated from data to update so will be impacted more by the batch size. Our result (figure \ref{fig:  figure 567 Impact of batchsize in DDN(4,16) weather}) exactly matched this speculation. We can observe that the NGD methods converge very faster compared to SGD methods when the batch size is high but the convergence significantly worsens when the batch size is reduced. Hence, the NGD methods are highly dependent on batch size.

\begin{figure}[h!]
\centering
\begin{subfigure}[b]{0.29\textwidth}
    \centering
    \includegraphics[width=\textwidth]{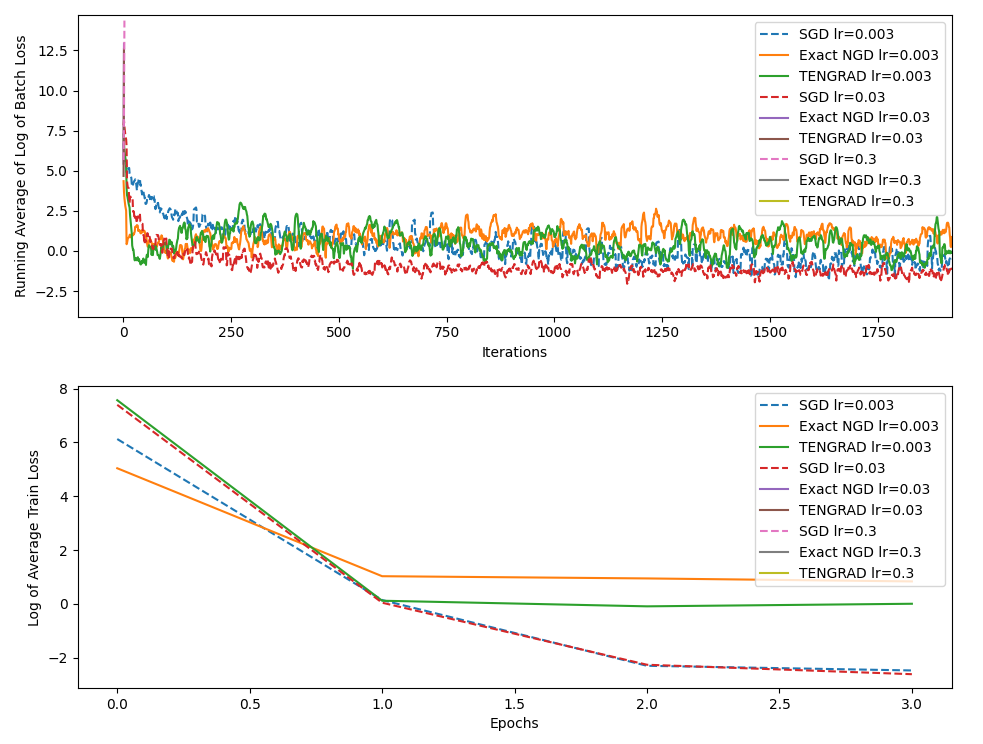}
    \caption{Batch Size=8}
    \label{fig: figure 5 batch size=8}
\end{subfigure}
\hfill
\begin{subfigure}[b]{0.29\textwidth}
    \centering
    \includegraphics[width=\textwidth]{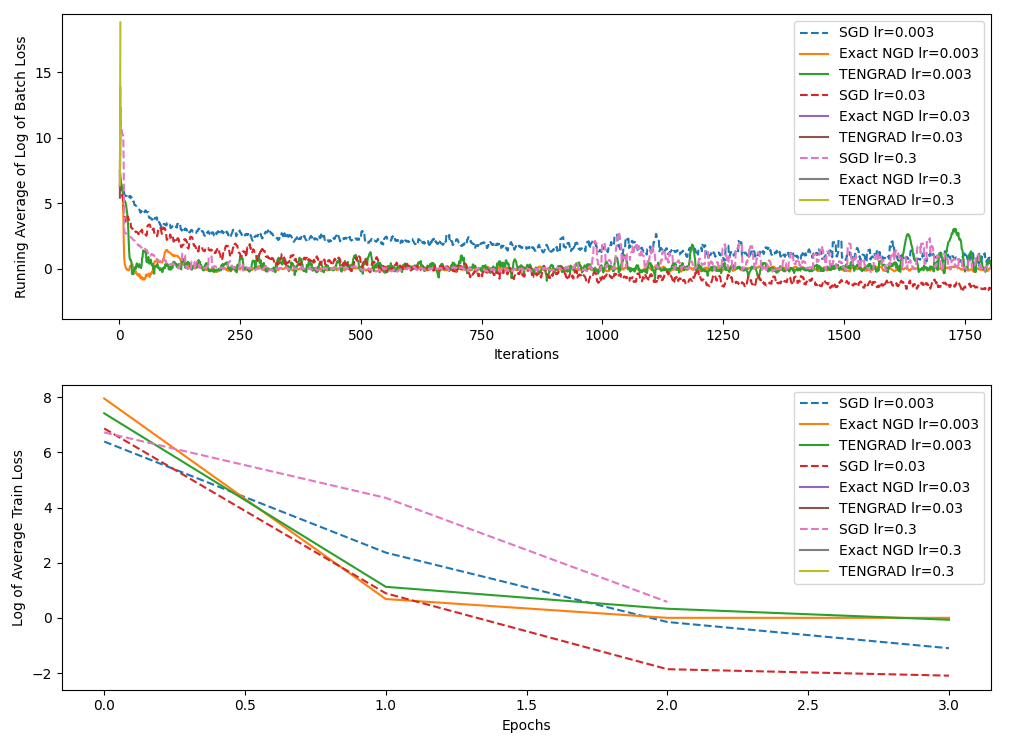}
    \caption{Batch Size=32}
    \label{fig: figure 6 batch size=32}
\end{subfigure}
\hfill
\begin{subfigure}[b]{0.29\textwidth}
    \centering
    \includegraphics[width=\textwidth]{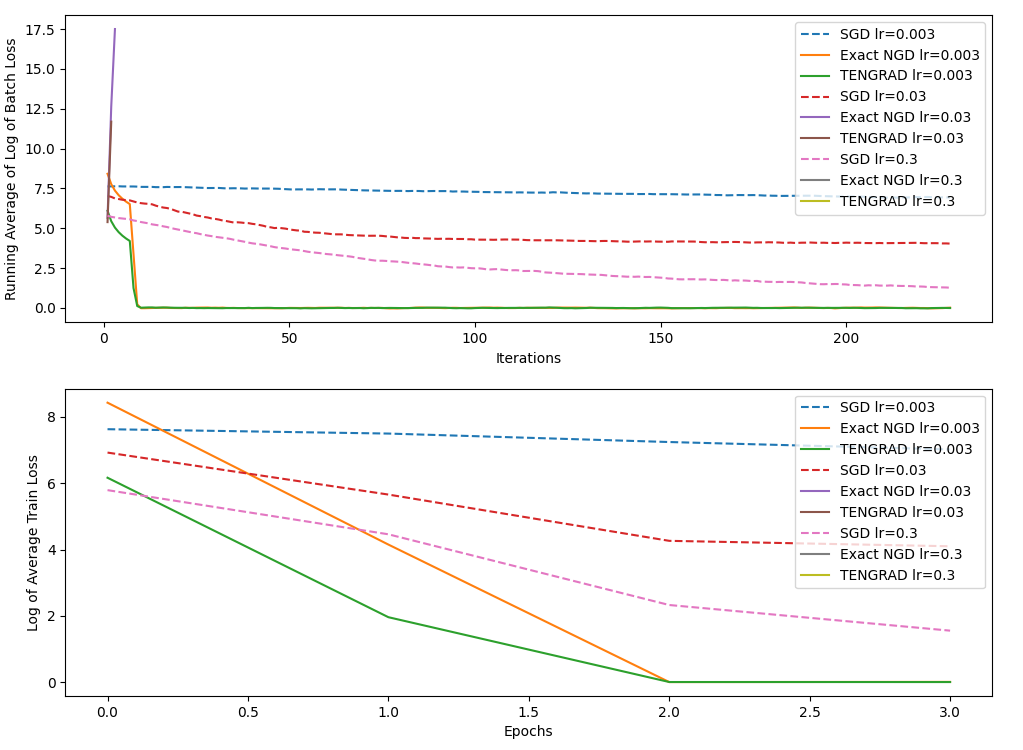}
    \caption{Batch Size=1024}
    \label{fig: figure 7 batch size=1024}
\end{subfigure}
    \caption{Impact of batch size in convergence. DNN with hidden layers=[32,16] in weather dataset}
    \label{fig:  figure 567 Impact of batchsize in DDN(4,16) weather}
\end{figure}

\subsection{Performance}
Figure \ref{fig:  figure 12 DDN(4) weather}, \ref{fig:  figure 34 Deep and/or Wide DDN(4) weather}, and \ref{fig:  figure 567 Impact of batchsize in DDN(4,16) weather} shows that in certain situations NGD methods often converge faster than SGD. However, does that translate to actual better performance in terms of the objective value at convergence i.e. convergence to a better or as good value as SGD method? In this subsection, I will try to answer these questions based on the experiments.

The argument \cite{10.1162/089976698300017746} that NGD methods perform better when the loss function can be well approximated by a convex quadratic aligns with the results that is obtained from the experiment. The first important factor is the batch size itself as discussed in subsection \ref{subsection: convergence}. The loss function is estimated with data of batch size that keeps changing based on the data sample itself and this variance is highly dependent on the sample size. From another perspective, the performance is dependent on the reliability of the empirical FIM matrix itself which is dependent on the batch size used to estimate it.  From figure \ref{fig:  figure 567 Impact of batchsize in DDN(4,16) weather}, it's evident that the loss at convergence for NGD methods is better for higher values of batch size but this performance degrades along with convergence with reduced batch size.  Another factor seems to be the complexity of the model. The performance seems to degrade as models get complex and as a result, the loss function curvature gets complex as seen in figure \ref{fig:  figure 12 DDN(4) weather} and \ref{fig:  figure 34 Deep and/or Wide DDN(4) weather}. Exact NGD methods are not feasible for deep neural networks, so we compare the loss at convergence for TENGraD and SGD methods. For this experiment, we use a batch size=128 that is around the value that is usually used in the deep learning community. We used three datasets for this: Weather \footnote{Weather dataset: https://www.kaggle.com/datasets/muthuj7/weather-dataset}(Regression), House Price\footnote{House Price dataset: https://www.kaggle.com/datasets/harlfoxem/housesalesprediction}(Regression) and Ecoli \footnote{Ecoli dataset: \cite{Dua:2019}}(Classification). TENGraD method seems to work better in the least square loss function compared to the cross-entropy loss function(Table \ref{table: performance of TENGraD and SGD} ).

\begin{table}
  \caption{Log of Training Loss at Convergence using batch size $m=128$}
  \label{table: performance of TENGraD and SGD}
  \centering
  \begin{tabular}{|l|l|l|l|l|}
    \toprule
    Hidden Layers & Method & Weather & House Price & Ecoli\\
    \midrule
    \multirow{2}{*}{4} & SGD & -0.45 &-0.192 & \textbf{-1.96}\\  
    & TENGraD & \textbf{-2.96} &\textbf{-1.120} & -0.16 \\ 
    \midrule
    \multirow{2}{*}{32,16}& SGD & -1 &-0.794 & \textbf{-1.37}\\ 
    & TENGraD & \textbf{-3.68} &\textbf{-0.876} & 0.59 \\ 
    \midrule
    \multirow{2}{*}{32,16,16} & SGD & 0.24&-0.16 & {0.39}\\ 
    & TENGraD & \textbf{-1.65} &\textbf{-1.06} & \textbf{0.25}\\ 
    
    \bottomrule
  \end{tabular}
\end{table}

\subsection{Stability}
The update using NGD is comparatively very noisy. The low batch size and the high number of parameters exacerbate this problem. It's very common for the FIM formed by the equation \ref{eqn: empirical FIM} to be ill-conditioned. It's very likely that the empirical FIM matrix has either very large eigenvalues or very small eigenvalues, both resulting in the instability of the NGD methods. From equation \ref{eqn: empirical FIM}, we can see that FIM is very likely to be rank deficient. Upon taking inverse of FIM, the inverse matrix either blows up or zeros the update direction obtained using natural gradient(equation \ref{eqn: natural gradient}). In addition, a stochastic empirical approximation of FIM makes it even worse. This is the reason why we added a constant $\beta I$ in equation \ref{eqn: whole NGD}. Adding a small value of $\beta$ prevents the condition of FIM being a singular matrix but the inverse would give a natural gradient having very huge matrices. However, if we add a bigger value of $\beta$, we can circumvent this problem, but this would defeat the purpose of using curvature-aware FIM and will be a scaled version of SGD itself for a high value of $\beta$. In the experiment, this instability issue seems to be more serious in a model trained for a classification task with cross-entropy loss function. We also found NGD methods to be very sensitive to the learning rate $\alpha$, so using a decaying value of it and a regularization in the weights of the network helped to stabilize the training of the NGD methods. Occasionally, the model could even diverge if continued training after converging to a fixed value. The decay learning rate also helped in this. 

\subsection{Scalability}
\begin{figure}
\begin{subfigure}[b]{0.48\textwidth}
    \centering
    \includegraphics[width=\textwidth]{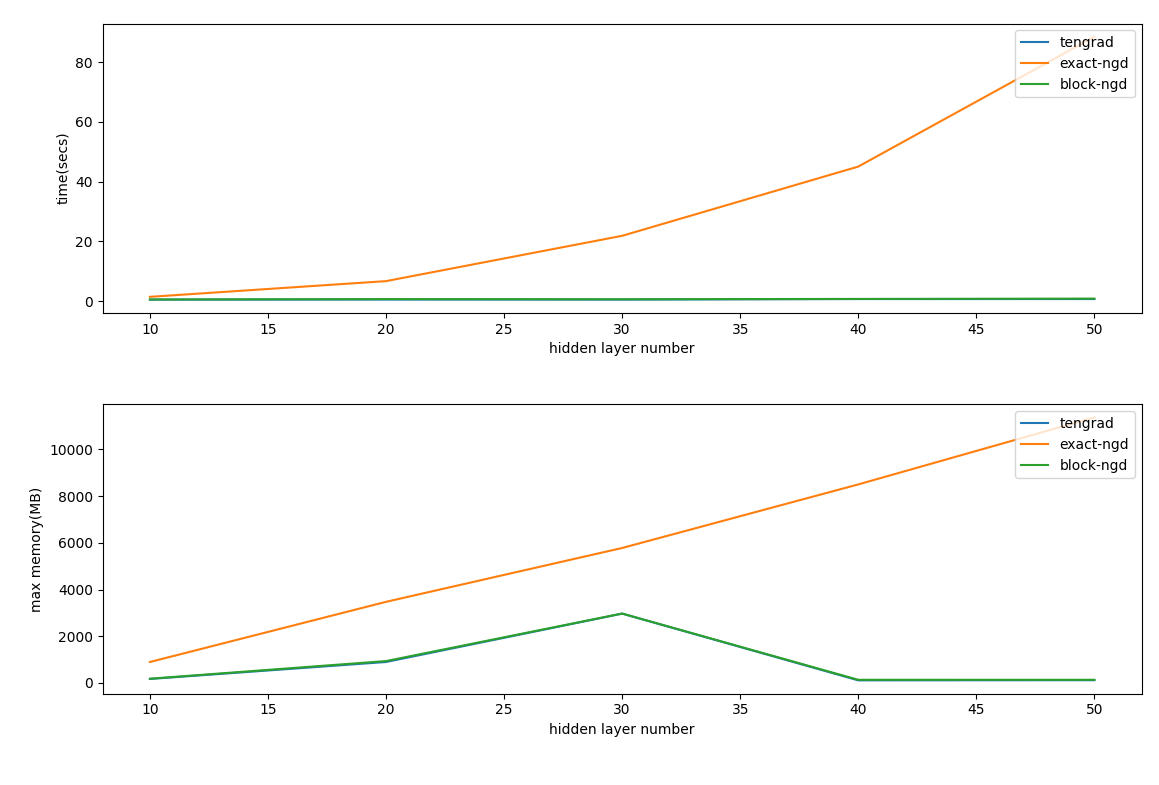}
    \caption{Time and memory requirement for Exact NGD, Blockwise NGD and TENGraD}
    \label{fig: figure 8 time and memory requirements}
\end{subfigure}
\hfill
\begin{subfigure}[b]{0.48\textwidth}
    \centering
    \includegraphics[width=\textwidth]{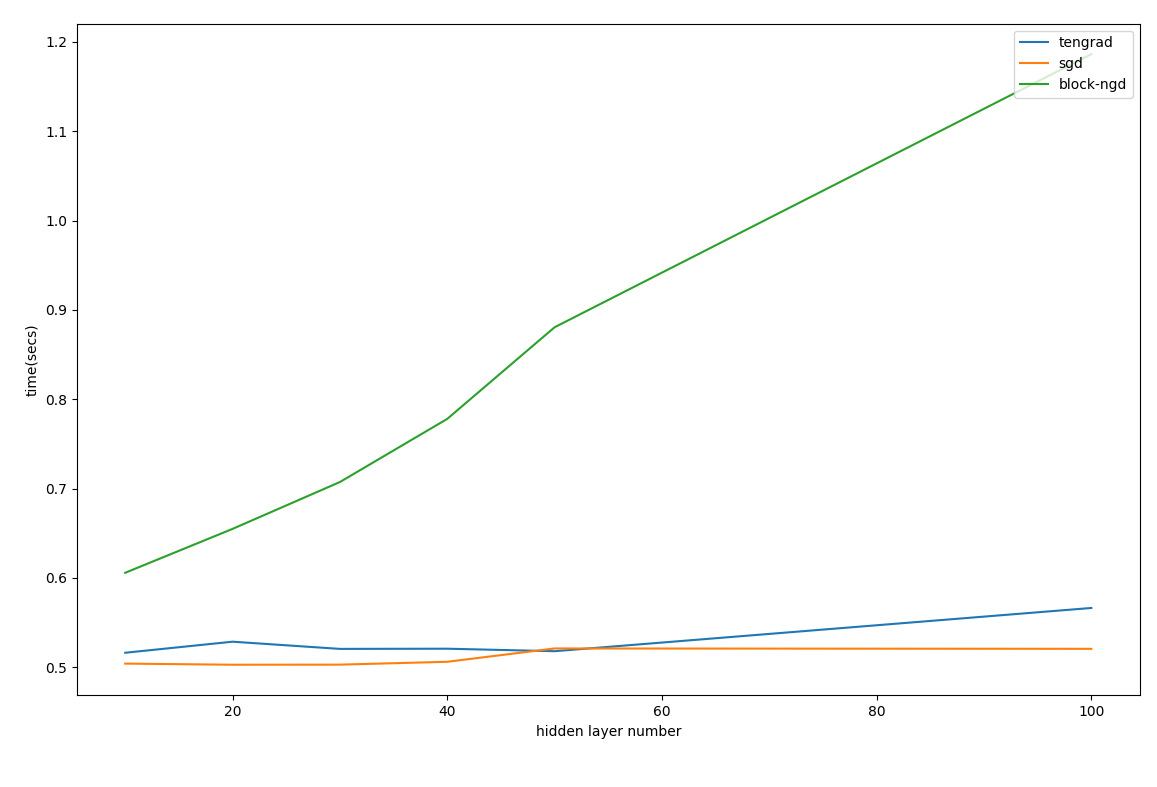}
    \caption{Time comparison between SGD, Blockwise NGD and TENGraD}
    \label{fig: figure 9 time requirement comparison}
\end{subfigure}
    \caption{Comparison of memory and time requirements in a model with each hidden layer has 20 nodes}
    \label{fig:  figure89 scalability}
\end{figure}

The NGD method in its original form scales $O(n^2)$ in space and $O(n^3)$ in time complexity if n is the total number of parameters of a model. The approximation methods alleviate this problem with certain assumptions. Blockwise NGD assumes the FIM to be block diagonal and  TENGraD further uses Woodbury Matrix Identity to change the inversion of the block-wise FIM matrix into consecutive time-efficient matrix operations. The efficiency in time and space requirements obtained from them in comparison to the original NGD can be clearly seen in figure \ref{fig: figure 8 time and memory requirements}. The time and space requirement for Exact NGD grows so fast making it infeasible for models with large parameters like deep neural network. The computation requirement in TENGraD is drastically reduced from the Blockwise NGD and scales in a similar fashion to the SGD method (figure \ref{fig: figure 9 time requirement comparison}).
\section{Conclusion}
With the new approximation method like TENGraD, the natural gradient method can easily scale even for deep learning models with a very high number of parameters. In certain conditions, NGD methods can definitely converge faster and reduce the number of iterations required. This will in turn lead to a reduction in time, as the time requirement for TENGraD scales similar to SGD. It also leads to better performance if the loss function can be approximated well enough by a convex quadrature. Theoretically, it has such an elegant interpretation as a steep descent in the distribution space, however, in practice, the stability issues and the hypersensitivity with respect to the hyperparameters make it less appealing.

The methods like RMSprop and Adam also maintain the running second-order moments approximating the FIM by considering it to be a diagonal matrix that makes the computation, storing, and inverting to O(n). They are relatively stable and work well with deep-learning models. I believe NGD can have a huge impact on deep learning and optimization but not with its current nascent state. Further research needs to be done on making NGD methods stable and perform at least as well as the established first-order methods. Adding momentum to the NGD methods might help and is reserved for future work.
\newpage
\bibliography{ref.bib}
\bibliographystyle{plainnat}
\newpage
\appendix
\section{Appendix\protect\footnote{The mathematical deviations in this section are adapted from the works of \cite{martens2020new, kristiadi}}}
\subsection{Expectation of score function} \label{appendix proof: expectation of score function}
\paragraph{Claim:}The expected value of the score function is 0.
\begin{align}
    \expectation_{\probability}[\scorefunction] &= \expectation_{\probability}[\nabla_{\theta}\loglikelihood] \nonumber\\
    &= \int_{\xsample} \probability \nabla_{\theta}\loglikelihood d\xsample \nonumber\\
    &= \int_{\xsample} \probability \frac{\nabla_{\theta}\probability}{\probability} d\xsample \nonumber\\
    &= \int_{\xsample} \nabla_{\theta}\probability d\xsample \nonumber\\
    &= \nabla_{\theta} \int_{\xsample} \probability d\xsample \nonumber\\
    &= \nabla_{\theta} 1 \nonumber\\
    \therefore \expectation_{\probability}[\scorefunction] &= 0 
\end{align}

\subsection{Relationship between Hessian and FIM } \label{appendix proof: relationship between Hessian and FIM}
The hessian($\hessian$) of the model $f(x|\theta)$ is the second-order partial derivative of the $\loglikelihood$. This is the same as the Jacobian($\jacobian$) of the gradient of $\loglikelihood$.
\begin{align}
    \hessian(\loglikelihood) &= \jacobian(\nabla_{\theta}\loglikelihood) \nonumber \\
    &= \jacobian(\frac{\nabla_{\theta}\probability}{\probability}) \nonumber \\
    &= \frac{\probability\hessian(\probability) - \nabla_{\theta}\probability \nabla_{\theta}{\probability}^T}{\probability^2} \nonumber \\
    &= \frac{\hessian(\probability}{\probability} - \frac{\nabla_{\theta}\probability \nabla_{\theta}{\probability}^T}{\probability^2} \nonumber \\
    &= \frac{\hessian(\probability}{\probability} - \nabla_{\theta}\loglikelihood \nabla_{\theta}{\loglikelihood}^T \\
    \expectation_{\probability}[\hessian(\loglikelihood)] &= \expectation_{\probability}\left[\frac{\hessian(\probability}{\probability}\right] - \expectation_{\probability}[\nabla_{\theta}\loglikelihood \nabla_{\theta}{\loglikelihood}^T]\nonumber\\
    &= \int_{\xsample} \probability \frac{\hessian(\probability}{\probability} d\xsample - \fim \nonumber \\
    &= \int_{\xsample} \hessian(\probability) d\xsample - \fim \nonumber \\
    &=  \hessian(\int_{\xsample}\probability d\xsample) - \fim \nonumber \\
    &=  \hessian(1) - \fim \nonumber \\
    &=  0 - \fim \nonumber \\
    \therefore \fim &= - \expectation_{\probability}[\hessian(\loglikelihood)] 
\end{align}

\subsection{Relationship between FIM and KL divergence } \label{appendix proof: FIM and KL divergence}
Let $\theta$ and $\theta'$ be two instances of the parameter of the model. Then, the KL divergence between the model's distribution corresponding to these two instantiations is $\text{KL}(p(x|\theta)||p(x|\theta'))$.

\begin{align}
    \text{KL}(p(x|\theta)||p(x|\theta')) &= \expectation_{p(x|\theta)}\left[log\left(\frac{p(x|\theta)} {p(x|\theta')}\right)\right] \nonumber \\
    &= \expectation_{p(x|\theta)}[log(p(x|\theta))] - \expectation_{p(x|\theta)}[log(p(x|\theta'))] \label{eqn: KL}\\
    \nabla_{\theta'}\text{KL}(p(x|\theta)||p(x|\theta')) &= \expectation_{p(x|\theta)}[\nabla_{\theta'}log(p(x|\theta))] - \expectation_{p(x|\theta)}[\nabla_{\theta'}log(p(x|\theta'))]\nonumber\\
    &= 0 - \expectation_{p(x|\theta)}[\nabla_{\theta'}log(p(x|\theta'))] \nonumber \\
    &= - \expectation_{p(x|\theta)}[\nabla_{\theta'}log(p(x|\theta'))] \label{eqn: first-order derivative of KL}\\
    \nabla_{\theta'}^2\text{KL}(p(x|\theta)||p(x|\theta')) &= - \expectation_{p(x|\theta)}[\nabla_{\theta'}^2log(p(x|\theta'))] \nonumber \\
    &= - \expectation_{p(x|\theta)}[\hessian(l(\theta'|x))] \label{eqn: second order derivative of KL}
\end{align}
Evaluating at $\theta'$ = $\theta$, then
\begin{equation}
    \nabla_{\theta'}^2\text{KL}(p(x|\theta)||p(x|\theta'))|_{\theta'=\theta} = - \expectation_{p(x|\theta)}[\hessian(l(\theta|x))] = \fim
\end{equation}

\subsection{NGD update equation using Woodbury Identity} \label{appendix proof: NGD Woodbury}
Assuming $W_l$ and $g_l$ to be vectors corresponding to the parameters of $l^{th}$ layer and their gradients,
\paragraph{NGD update}
\begin{equation}
    W_l(k+1) = W_l(k) - \alpha[F_l(W_l(k)) + \beta I]^{-1}g_l(k),  \forall_{l=1,....,L} \label{appendix eqn: ngd update}
\end{equation}
\paragraph{Woodbury Matrix Identity}
\begin{equation}
    (A + UCV)^{-1} = A^{-1} - A^{-1}U(C^{-1} + VA^{-1}U)^{-1}VA^{-1} \label{appendix eqn: woodbury matrix identity}
\end{equation}

Substituting $A = \beta I$, $C=I$, $U=\frac{J_l(k)^T}{m}$ and $V=J_l(k)$ in equation \ref{appendix eqn: woodbury matrix identity}, we get
\begin{align}
    & \left(\beta I + \frac{J_l(k)^T}{m} I J_l(k)\right)^{-1} \nonumber \\
    &= \left(\beta I + \frac{J_l(k)^T J_l(k)} {m}\right)^{-1} \nonumber \\
     &= {(\beta I)}^{-1} - {(\beta I)}^{-1}\frac{J_l(k)^T}{m}\left(I^{-1} + J_l(k){(\beta I)}^{-1}\frac{J_l(k)^T}{m}\right)^{-1}J_l(k){(\beta I)}^{-1} \nonumber \\
    &=  \frac{I}{\beta} - \frac{I}{\beta}\frac{J_l(k)^T}{m}\left(I + J_l(k)\frac{I}{\beta}\frac{J_l(k)^T}{m}\right)^{-1}J_l(k)\frac{I}{\beta}\nonumber\\
    &=  \frac{I}{\beta} - \frac{J_l(k)^T}{m\beta}\left(I + \frac{J_l(k) J_l(k)^T}{m\beta}\right)^{-1}\frac{J_l(k)}{\beta}\nonumber\\
    &=  \frac{I}{\beta} - \frac{J_l(k)^T}{m\beta}\beta\left(\beta I +\frac{ J_l(k) J_l(k)^T}{m}\right)^{-1}\frac{J_l(k)}{\beta}\nonumber\\
    &=  \frac{I}{\beta} - \frac{J_l(k)^T}{m}\left(\beta I +\frac{ J_l(k) J_l(k)^T}{m}\right)^{-1}\frac{J_l(k)}{\beta}\nonumber
    \end{align}
    \begin{align}
    \left(\frac{J_l(k)^T J_l(k)} {m} + \beta I\right)^{-1} &= \frac{1}{\beta}\left(I - \frac{J_l(k)^T}{m}\left(\beta I +\frac{ J_l(k) J_l(k)^T}{m}\right)^{-1}J_l(k)\right)\nonumber
    \end{align}
    \begin{align}
    \therefore W_l(k+1) &= W_l(k) - \alpha\frac{1}{\beta}\left(I - \frac{J_l(k)^T}{m}\left(\beta I +\frac{ J_l(k) J_l(k)^T}{m}\right)^{-1}J_l(k)\right)g_l(k) \nonumber \\
    & = W_l(k) - \frac{\alpha}{\beta}\left(g_l(k) - \frac{J_l(k)^T}{m}\left(\beta I +\frac{ J_l(k) J_l(k)^T}{m}\right)^{-1}J_l(k)g_l(k)\right)
\end{align}
\subsection{TENGraD} \label{tengrad description}
This model was proposed by \cite{soori2021tengrad} and the details in the section are based on the original paper.\\
\emph{Notation}:\\
$\odot$: Hadamard product between two matrices i.e. elementwise product\\
$\otimes$: Kronecker product\\
$||.||_2$: $l_2$ norm or euclidean norm\\
$\ast$: column-wise Khatri-Rao product\\
$M:i$: $i^{th}$ column of matrix M

A deep neural network with $L$ layers is a functional approximator in the form of $f(x, W)$ where x is the input and W are the parameters of the network. Let $n$ be the total number of data samples and $m$ be the batch size that is used to train the model. Then, for each layer $l$, let the input size be $d_i^l$ and the output size be $d_o^l$. The weight of the layer l is denoted by $W_l\in \mathbb{R}^{d_i^l\times d_o^l}$, the input by $I_l\in\mathbb{R}^{d_i^l\times m}$ and the preactivation output by $O_l\in\mathbb{R}^{d_o\times m}$.
\begin{equation}
    O_l = W_l^T I_l
\end{equation}
With the sample $(x,y)$ of size $n$, the model is trained to learn the parameters of the model that explain the data. This is usually done by minimizing an average loss function on the training data with respect to the model's parameters (equation \ref{eqn: average loss function}). Usually, $x$ is referred to as features and $y$ as labels.
\begin{equation}
    \mathcal{L}(W) = \frac{1}{n}\sum_{i=1}^n l(f(x_i,W),y) \label{eqn: average loss function}
\end{equation}
Here, $W = [vec(W_1)^T, vec(W_2)^T, ...., vec(W_L)^T]$ denotes the vectorized form of all the parameters of the model.  let $p$ represents the total count of parameters i.e $W\in\mathbb{R}^{p}$.

\paragraph{Gradient and Jacobian:}
The gradient $g$ of the loss function (equation \ref{eqn: average loss function}) with respect to $W$ is calculated over the batch data and often used for the training. This is the partial derivative of the $\mathcal{L}$ with $W$. Similarly, let $g_l$ denote the gradient corresponding to the $l^{th}$ layer.
\begin{align}
    g &= \nabla_{W}\mathcal{L}(W) = \frac{1}{m} \sum_{i=1}^{m}\nabla_W l(f(x_i,W),y_i) \in \mathbb{R}^{p} \\
\end{align}

Backpropagation is used while training the neural networks during which each layer receives the partial derivatives of the loss with respect to its outputs for each data in the batch. Let's denote this derivative of the loss with respect to the preactivation output by $G_l\in\mathbb{R}^{d_o^l\times m}$. $G_l[i,j]$ denotes the partial derivative of loss with respect to $i^{th}$ preactivation output of layer $l$ for $j^{th}$ data in the batch. Then, the gradient of the loss with respect to the parameters of the layer ($g_l$) is given by
\begin{equation}
    g_l = \frac{1}{m} I_l G_l^T \label{eqn: gradient for layer l}
\end{equation}
For each layer $l$, let $J_l$ denote the Jacobian of the loss with respect to each layer parameter for $m$ data in a batch. These jacobians can be concatenated to form the Jacobian$J$ with respect to all the parameters of the model.
\begin{align}
    J_l = (I_l \ast G_l )^T \in \mathbb{R}^{m\times d_i^l d_o^l} \label{eqn: layer jacobian}\\
    J = [J_1, J_2, ....., J_L] \in \mathbb{R}^{m \times p}
\end{align}

\subsubsection{Parameter Updates}
Here, we will see how the updates in TENGraD is derived from the original NGD for efficiency and scalability purpose.
\paragraph{Original NGD Update at iteration k for DNN:}
\begin{equation}
    vec(W(k+1)) = vec(W(k)) - \alpha[F(W(k)) + \beta I]^{-1}vec(g(k)) \label{eqn: whole NGD}
\end{equation}
where $\alpha$ is the learning rate, $F$ is the FIM. A non-negative damping factor $\beta$ is added to prevent the singular case of F when the model is overparameterized to ensure the inversion of the matrix in equation \ref{eqn: whole NGD}. The FIM  coincides with the Gauss-Newton matrix when the output conditional distribution is assumed to be an exponential distribution.
\begin{align}
    F(W(k)) &= \mathbb{E}_{p(x,y)}[\nabla_W logp_W(y|x)\nabla_W logp_W(y|x)^T] \nonumber\\
    &\approx \frac{1}{m} \sum_{i=1}^m \nabla_{W}l(f(x_i,W(k),y_i) \nabla_{W}l(f(x_i,W(k),y_i)^T && \text{(Using Empirical FIM)}\nonumber\\
    &= \frac{1}{m} \sum_{i=1}^m \nabla_{W}l_i \nabla_{W}l_i^T\nonumber\\
    &= \frac{1}{m} J(k)^T J(k)
\end{align}

 \paragraph{NGD Update with Block diagonal FIM:}
 When the FIM is approximated using a block diagonal matrix, then each of layer can be updated separately using FIM corresponding to each layer ($F_l$). The update for layer $l$ becomes
 \begin{align}
    F_l(W_l(k)) &= \frac{1}{m}J_l(k)^T J_l(k)\label{eqn: block FIM}\\
    W_l(k+1) &= W_l(k) - \alpha[F_l(W_l(k)) + \beta I]^{-1}g_l(k),  \forall_{l=1,....,L}\label{eqn: block diagonal NGD}
 \end{align}

\paragraph{Time Efficient Exact FIM inversion:}
The inversion is the most computationally intensive operation in the equation \ref{eqn: block diagonal NGD}. TENGrad computes the exact FIM inverse matrix for all the layers by using factorization and reusing intermediate values in the computation. Using Woodbury identity for each block, update changes into equation \ref{eqn: Update equation Using Woodbury Identity}(proof in Appendix \ref{appendix proof: NGD Woodbury}).
\begin{multline}
    vec(W_l(k+1)) = vec(W_l(k)) - \\ \frac{\alpha}{\beta}\left(vec(g_l(k)) - \underbrace{\frac{J_l(k)^T}{m}}_{C}\underbrace{\left(\frac{J_l(k) J_l(k)^T)}  {m}+\beta I\right)^{-1}}_{A}\underbrace{J_l(k)vec(g_l(k))}_{B}\right) \label{eqn: Update equation Using Woodbury Identity}
\end{multline}
All the components in the equation \ref{eqn: Update equation Using Woodbury Identity} are computed efficiently in TENGraD. The gradient of layer $l$ from the data $i$ is given by $g_{l,i} = I_{l,:i}G_{l,:i}^T$(equation \ref{eqn: gradient for layer l}) which is a rank one matrix obtained by outer product.   Then $[J_lJ_l^T]_{i,j}=<vec(g_{l,i}), vec(g_{l,j})>$. We know that for the outer product $vec(uv^T)=u \otimes v$ which makes $[J_lJ_l^T]_{i,j} = <I_{l,:i} \otimes G_{l,:i}, I_{l,:j} \otimes G_{l,:j}>$. Using $<u_1 \otimes v_1, u_2 \otimes v_2>$ $= u_1^Tu_2.v_1^Tv_2$, we get $[J_lJ_l^T]_{i,j} = I_{l,:i}^TI_{l,:j}.G_{l,:i}G_{l,:j}$. Extending this to the batch of size $m$, we have  
\begin{equation}
    J_lJ_l^T=(I_l \ast G_l )^T (I_l \ast G_l) = I_l^T I_l \odot G_l^T G_l= C_1 \odot C_2  \label{eqn: step involving A}
\end{equation}
The gram Jacobian is the Hadamard product of two small matrices of size $m\times m$. It can be efficiently computed with matrix product and without having to compute the jacobian. Part A in equation \ref{eqn: Update equation Using Woodbury Identity} involves small matrices and can be computed efficiently.
Part B and C involve the use of jacobian and requires $m$ times the parameter size memory requirement to store it. TENGraD circumvents this problem by storing small data structures and  recomputing the jacobian as per necessity. Using equation \ref{eqn: layer jacobian} and properties of column-wise Khatri-Rao product, the C part in equation \ref{eqn: Update equation Using Woodbury Identity} can be computed as
\begin{equation}
    J_lvec(g_k) = (I_l \ast G_l)^T vec(g_l(k)) = ((g(k)^TI) \odot G)1 \label{eqn: step involving B}
\end{equation}
This involves two efficient matrix operations without having to form the jacobian. Then, the product of A and B term results in a vector $v\in\mathbb{R}^m$ requiring $J_l(K)^T.v$ as the final operation which can also be done efficiently.
\begin{equation}
    J_l^T = (I \ast G)v = I(v1^T \odot G^T) \label{eqn: step involving C}
\end{equation}
With the use of the step in equation \ref{eqn: step involving A} for computing part A, equation \ref{eqn: step involving B} for computing part B, and equation \ref{eqn: step involving C} for part C, TENGraD computes the updates in equation \ref{eqn: Update equation Using Woodbury Identity} very efficiently in terms of time and space requirements.

\end{document}